\documentclass[11pt]{article}

\usepackage[preprint]{acl}

\usepackage{xspace}
\makeatletter
\ifacl@anonymize
  \newcommand{\company}{[Company]\xspace}
\else
  \newcommand{\company}{LinkedIn\xspace}
\fi
\makeatother

\usepackage{times}
\usepackage{latexsym}

\usepackage[T1]{fontenc}

\usepackage[utf8]{inputenc}

\usepackage{microtype}

\usepackage{inconsolata}

\usepackage{graphicx}

\usepackage{tikz}
\usepackage{makecell}
\usepackage{algorithm}
\usepackage{algpseudocode}
\usepackage{tablefootnote}
\usepackage{caption}
\usepackage{booktabs}
\usepackage{threeparttable}
\usetikzlibrary{positioning,arrows.meta,shapes.geometric,fit,calc,shapes.misc}

\title{Self-evolving Agentic Customer Support System at \company}

\author{
 Chih Hui Wang\thanks{Equal contribution}, Mengdie Tu\footnotemark[1], Qianyun Zhang\footnotemark[1], Wei Wu\footnotemark[1],
\\
 {\bf Lili Zhou, Mingqi Shen, Changshuai Wei\thanks{Corresponding author}}
\\
LinkedIn
\\
\texttt{\{ciwang, metu, qazhang, wwu1, lizhou, minshen, chawei\}@linkedin.com}
}

\begin{document}
\maketitle
\begin{abstract}
Enterprise support agents operate in rapidly changing environments where policies, product capabilities, and knowledge bases evolve continuously, making static assistants brittle and costly to maintain. We present \company's self-evolving agentic support system, which integrates retrieval-augmented generation with evolutionary auto-prompting and a modular, production-aligned evaluation framework to enable safe, continuous improvement without retraining foundation models. The system treats prompts, retrieval, and evaluation as a closed-loop, versioned workflow with operational guardrails. Offline simulations and ablations show clear quality gains over vanilla RAG and baseline agents, including reduced hallucinations and improved response completeness. In a two-week user-randomized A/B test on \company's production support traffic, the integrated self-evolved workflow increased QA self-serve by 9.0 percentage points, cancellation self-serve by 4.8 points, and routing accuracy by 30.6 points. These results demonstrate a practical path to scalable, self-evolving AI agents in real-world enterprise settings.
\end{abstract}

\section{Introduction}
Enterprise customer support is increasingly mediated by AI agents, yet real-world deployments face a reliability problem qualitatively different from standard chatbot benchmarks. At \company, support spans many product surfaces, multiple lines of business, and dozens of languages, while the underlying ecosystem changes continuously: product launches reshape the knowledge base, content is published or deprecated, retrieval indices are refreshed, prompts evolve, and underlying LLMs and tool APIs change over time. Even small upstream changes can induce regressions---hallucinations, stale guidance, broken tool flows, or inconsistent multilingual tone---making a ``set-and-forget'' assistant brittle and expensive to maintain. Traditional production support agents---built on a handcrafted prompt, a fixed retrieval pipeline, and periodic human QA---do not scale with this velocity, and lack a principled feedback loop that can detect regressions, localize failures to specific components, and adapt without heavy human intervention.

Three lines of recent work motivate the design we present. \emph{Prompt optimization} treats prompts as programs, with score-based search (APE \cite{zhou2023ape}), LLM-as-optimizer (OPRO \cite{yang2024opro}), evolutionary search (PromptBreeder, EvoPrompt \cite{fernando2024promptbreeder,guo2024evoprompt}), and declarative compilation (DSPy \cite{khattab2023dspy}) all showing that prompts can be evolved offline against task metrics and shipped as versioned artifacts. \emph{LLM-as-a-judge} evaluation enables scalable assessment of multi-dimensional quality \cite{zheng2023mtbench,liu2024geval}, with documented biases that can be mitigated through structured rubrics, length control \cite{dubois2024lengthcontrolled}, and open evaluator models \cite{kim2024prometheus2}. \emph{Retrieval-augmented generation} grounds responses in proprietary knowledge \cite{lewis2020rag,karpukhin2020dpr,gao2023hyde}, and recent self-reflective variants (Self-RAG, CRAG \cite{asai2024selfrag,yan2024crag}) treat retrieval as an explicit agent action rather than a preprocessing step, aligning with reasoning-and-acting paradigms \cite{yao2023react}.

For enterprise support, the operational implication of these advances is that prompts, retrievers, and evaluators all become \emph{versioned artifacts} that can change frequently---and often function like policy changes affecting routing, escalation thresholds, safety posture, and brand tone across multiple products and languages. A self-evolving system therefore requires not just optimization, but the engineering discipline to deploy these changes safely: regression testing, staged rollout, gated promotion, and rollback. Our evaluation layer must in turn be reliable enough to drive these decisions at scale, robust to knowledge drift, and decomposed into interpretable signals so failures can be attributed and acted on.

We co-evolve the three layers under strict version control, deliberately avoiding fine-tuning because support knowledge changes too quickly for retraining cycles and because retrieval-based knowledge injection often outperforms unsupervised fine-tuning for factual updates \cite{ovadia2024finetuneorretrieve,soudani2024finetuningvsrag}. Our evaluation layer is structured as specialized evaluator agents \cite{wu2023autogen}, and our operational ``memory'' is implemented as bounded, auditable, versioned artifacts rather than unconstrained long-horizon memory \cite{packer2023memgpt,shinn2023reflexion}.

\textbf{Contributions.}
We (1) describe an end-to-end self-evolving support agent that remains reliable under continual upstream change via closed-loop evaluation and optimization; (2) present an automatic prompt evolution engine suitable for enterprise constraints (tone, policy, multilingual); (3) propose a modular, multi-signal evaluation framework actionable for debugging at scale; and (4) detail a version-controlled RAG-and-tools layer supporting reproducibility, safe rollout, and rollback.

\section{Methodology}
\subsection{Overall system}
Figure~\ref{fig:self_evolving_loop} shows the Support AI Agent as a closed-loop, self-evolving system in which prompting, retrieval, and evaluation form an explicit feedback cycle. At inference time the agent is conditioned by a system prompt from the Automatic Prompt Engine that encodes task instructions, tone, and implicit policies for reasoning depth and tool use. RAG is exposed as an explicit action rather than a preprocessing step, letting the agent decide when to retrieve and how to integrate evidence, consistent with reasoning-and-acting paradigms~\cite{yao2023react,schick2023toolformer}. Outputs are scored by a modular evaluation framework that decomposes quality into grounding, intent alignment, multilingual fidelity, and stylistic compliance using rule-based diagnostics and LLM-based judges~\cite{zheng2023mtbench}; aggregate fitness signals feed back into the prompt engine, which refines the system prompt via evolutionary search. The architecture exposes two coupled loops---an inner inference loop (agent~$\leftrightarrow$~RAG~$\leftrightarrow$~content lake) and an outer optimization loop (auto-prompt~$\rightarrow$~agent~$\rightarrow$~evaluator~$\rightarrow$~auto-prompt)---enabling scalable self-improvement and rapid adaptation across products and languages without modifying the base model or retrieval corpus.

\begin{figure*}[t]
\centering
\scalebox{0.85}{
\begin{tikzpicture}[
  font=\small,
  node distance=15mm and 32mm,
  box/.style={draw, rounded corners, align=center, inner sep=6pt, minimum width=3.5cm, minimum height=1.2cm},
  smallbox/.style={draw, rounded corners, align=center, inner sep=5pt, minimum width=3.2cm, minimum height=0.9cm},
  dashedbox/.style={draw, rounded corners, dashed, align=center, inner sep=6pt},
  arrow/.style={-{Stealth[length=2.2mm]}, line width=0.7pt},
  dashedarrow/.style={-{Stealth[length=2.2mm]}, dashed, line width=0.7pt},
  edgelabel/.style={font=\scriptsize\itshape, align=center}
]

\node[box] (agent) {Support AI Agent\\{\footnotesize (reason + act)}};
\node[box, left=of agent] (auto) {Auto-Prompt\\Engine\\{\footnotesize (evolutionary search)}};
\node[box, right=of agent] (eval) {Modular Evaluation\\Framework\\{\footnotesize (scores \& diagnostics)}};

\node[smallbox, below=22mm of agent] (rag) {RAG Tool\\{\footnotesize (retrieve evidence)}};
\node[smallbox, below=12mm of rag] (kb) {Versioned Content Lake\\{\footnotesize (help/learn/docs)}};

\node[dashedbox, above=14mm of agent] (user) {User Query\\+ Chat Context};
\node[dashedbox, above=14mm of eval] (out) {Agent Response\\(+ citations)};

\draw[arrow] (auto) -- node[edgelabel, above, yshift=2pt] {optimized\\system prompt} (agent);
\draw[arrow] (agent) -- node[edgelabel, above, yshift=2pt] {responses + traces} (eval);

\draw[arrow] (eval.south) -- ++(0,-1.4) -| node[edgelabel, pos=0.25, below, yshift=-2pt] {fitness signals (multi-metric)} (auto.south);

\draw[dashedarrow] (agent.285) -- node[edgelabel, right, xshift=4pt] {tool invocation\\(as action)} (rag.75);
\draw[dashedarrow] (rag.105) -- node[edgelabel, left, xshift=-4pt] {grounding docs\\+ provenance} (agent.255);
\draw[arrow] (rag) -- node[edgelabel, right, xshift=4pt] {retrieve top-$k$\\evidence} (kb);

\draw[arrow] (user) -- node[edgelabel, right, xshift=-2pt, yshift=-8] {request} (agent);
\draw[arrow] (eval.north) -- node[edgelabel, right, xshift=3pt, yshift=-10pt] {quality report\\\& failure types} (out.south);

\node[
  draw, rounded corners,
  inner sep=18pt,
  fit=(auto)(agent)(eval),
  label={[edgelabel, yshift=-2pt, xshift=2pt]above:Closed-loop self-evolution}
] (loop) {};

\node[align=left, font=\footnotesize, below=6mm of kb] (note) {%
\textbf{What evolves:} prompt policies (tool use, tone, structure)\\
\textbf{What stays fixed per iteration:} base LLM\textsuperscript{1} + content snapshot
};

\end{tikzpicture}
}
\caption{Self-evolving Support AI Agent architecture.}
\label{fig:self_evolving_loop}
\end{figure*}

\subsection{Auto-Prompt}
Manual prompt development is a major bottleneck in production: in our deployment, a single line-of-business iteration can take weeks. We replace this with an Automatic Prompt Engineering Engine (``Auto-Prompt'') that formulates prompt optimization as evolutionary search inspired by genetic algorithms~\cite{holland1992adaptation,whitley1994genetic}. Domain-specific business rules guide an LLM-driven initialization stage, after which a population of candidate prompts evolves through selection, crossover, and mutation, evaluated each generation against task-specific metrics; only the strongest survive into the next iteration. The full pseudocode is given in Algorithm~\ref{alg:genetic_prompt_optimization}.

Evolutionary search fits this setting because prompt optimization is black-box, non-differentiable search over discrete text, with fitness supplied by LLM and rule-based evaluators. Unlike single-candidate hill climbing, a population preserves diverse combinations of tone, tool-use policy, reasoning depth, and response structure; crossover recombines complementary instructions discovered in different lineages, while mutation provides local exploration. APE and OPRO optimize prompts through score-guided generation~\cite{zhou2023ape,yang2024opro}, and DSPy compiles declarative programs against metrics~\cite{khattab2023dspy}; our use of constrained evolution emphasizes population diversity, recombination, and immutable enterprise-policy filters. The ablation in Table~\ref{tab:auto_prompt_simulation} tests whether both evolutionary operators are necessary.

To prevent evolutionary drift from violating safety or policy requirements, business rules are treated as immutable hard constraints rather than optimization objectives: violating prompts are filtered or assigned zero fitness prior to selection. Crossover and mutation therefore operate within a constrained prompt space that preserves required behaviors by construction, ensuring evolutionary gains target accuracy without eroding compliance or product invariants.

\begin{algorithm}[h]
\caption{Genetic Prompt Optimization}
\label{alg:genetic_prompt_optimization}
\begin{algorithmic}[1]

\Require
Business rules $\mathcal{R}$,
evaluation dataset $\mathcal{D}$,
population size $N$,
elite size $K$,
maximum generations $G$,
scoring function $\mathcal{S}(\cdot)$

\Ensure Optimized prompt $p^*$

\State \textbf{Initialize} prompt population $P_0 \gets \textsc{LLMGenerate}(\mathcal{R}, N)$
\State Initialize evaluation cache $\mathcal{C} \gets \emptyset$

\For{$g = 1$ \textbf{to} $G$}

    \For{\textbf{each} prompt $p \in P_{g-1}$}
        \If{$p \notin \mathcal{C}$}
            \State Run LLM inference on $\mathcal{D}$ using $p$
            \State Compute fitness $f(p) \gets \mathcal{S}(p, \mathcal{D})$
            \State Store $(p, f(p))$ in cache $\mathcal{C}$
        \Else
            \State Retrieve $f(p)$ from $\mathcal{C}$
        \EndIf
    \EndFor

    \State Select top-$K$ elite prompts $E_g \subset P_{g-1}$ by fitness
    \State Initialize new population $P_g \gets E_g$

    \While{$|P_g| < N$}
        \State Sample parents $(p_i, p_j)$ from $E_g$
        \State $p' \gets \textsc{SemanticBlend}(p_i, p_j)$\footnote{SemanticBlend: LLM-based crossover that merges two prompts while preserving meaning and tone.}
        \State $p'' \gets \textsc{Mutate}(p')$\footnote{Mutate: small random edits to increase diversity and exploration.}
        \State Add $p''$ to $P_g$
    \EndWhile

\EndFor

\State \Return $p^* \gets \arg\max_{p \in P_G} f(p)$

\end{algorithmic}
\end{algorithm}

\subsection{Retrieval Augmented Generation (RAG)}
Support knowledge is proprietary and changes constantly, so we ground responses in retrieved content rather than parametric model knowledge~\cite{lewis2020rag,guu2020realm}. We depart from fixed retrieval pipelines and expose RAG as a \emph{tool} the agent invokes within its reasoning loop~\cite{yao2023react,schick2023toolformer}, letting the model decide when to retrieve and how to query, mirroring recent agentic RAG systems~\cite{press2023iterative,shinn2023reflexion}.

Our unified knowledge base consolidates three content sources: help articles, learning content, and product documentation microsites. They are refreshed by an ETL (Extract--Transform--Load) pipeline that chunks, embeds, facets, and retires content before loading it into the versioned content lake. Retirement is enforced through a snapshot pointer, so each refresh scopes retrieval to the latest snapshot and supersedes prior content without deleting it. Each document is tagged with three facets---\emph{product} (line of business), \emph{locale} (7+ languages), and \emph{snapshot date}---enabling scoped, reproducible retrieval~\cite{lin2021pretrained,gao2023rag}. At query time, hybrid dense--sparse retrieval is followed by semantic re-ranking~\cite{microsoft2023hybrid,nogueira2019passage}, returning ranked grounding documents with provenance (URL, locale, source) for citation and audit.\footnote{GPT-4o-mini (via Azure OpenAI service) was used as the base LLM in production at the time of writing.}

\subsection{Evaluation Framework}
\label{sec:evaluation-framework}
Operating across 36+ languages and many product lines with a constantly drifting knowledge base, we cannot rely on fixed gold sets or static benchmarks~\cite{lewis2019mlqa,longpre2021mkqa}. We instead use a modular, agent-aware framework that decomposes quality into interpretable dimensions---grounding, intent understanding, tool-execution correctness, and multilingual fidelity---each scored by specialized evaluators combining rule-based checks with LLM judgments. Multilingual quality is assessed by dedicated modules that detect untranslated spans, verify terminology, and measure fluency via XGLM perplexity~\cite{lin2022xglm}, while an LLM evaluator scores semantic fidelity under realistic dialog conditions. Production monitoring spans all supported locales, while the controlled benchmark in Section~\ref{sec:translation-eval} focuses on English$\leftrightarrow$Chinese as a challenging distant-language pair.

Unlike static benchmarks, evaluation is conditioned on the context retrieved at inference time, addressing knowledge drift directly~\cite{zheng2023mtbench}. Signals are aggregated through a consensus layer calibrated against periodic human annotations, stored in a versioned replayable format, and consumed by upstream optimization (auto-prompt and RAG iteration), closing the loop between agent behavior, measurement, and controlled self-evolution.

\section{Engineering Architecture and Deployment}
\label{sec:arch}

\begin{figure*}[t]
\centering
\resizebox{\textwidth}{!}{%
\begin{tikzpicture}[
  font=\small,
  box/.style={draw, rounded corners, align=center, minimum width=3.8cm, minimum height=0.9cm},
  bigbox/.style={draw, rounded corners, align=left, inner sep=6pt},
  arrow/.style={-{Stealth[length=2.2mm]}, line width=0.7pt},
  dashedarrow/.style={-{Stealth[length=2.2mm]}, dashed, line width=0.7pt}
]

\node[box] (ui) {User Chat Interfaces};
\node[box, below=8mm of ui] (orch) {Orchestration \&\\Fallback};

\node[bigbox, below=8mm of orch, minimum width=8.4cm] (agent) {
\textbf{Agent Runtime}\\
\quad Prompt assembly\\
\quad Context / RAG\\
\quad Tool invocation
};

\node[bigbox, right=10mm of agent, minimum width=6.2cm] (ext) {
\textbf{Versioned External Capabilities}\\
\quad LLM gateway\\
\quad Tool APIs\\
\quad Knowledge store\\
\quad Moderation
};

\node[bigbox, below=18mm of agent, minimum width=6.2cm] (telemetry) {
\textbf{Telemetry \& Event Stream}\\
\quad Traces\\
\quad Events\\
\quad ETL
};

\node[bigbox, right=10mm of telemetry, minimum width=7.0cm] (eval) {
\textbf{Modular Evaluation Agents}\\
\quad Groundedness\\
\quad Relevance\\
\quad Completeness\\
\quad Translation
};

\node[bigbox, right=10mm of eval, minimum width=7.2cm] (registry) {
\textbf{Versioned Artifacts Registry}\\
\quad Scores \\
\quad Datasets\\
\quad Regressions
};

\node[bigbox, below=10mm of eval, minimum width=10.2cm] (opt) {
\textbf{Optimization \& Safe Rollout}\\
\quad Prompt evolution\\
\quad Retrieval tuning\\
\quad Gated deploy / rollback
};

\draw[arrow] (ui) -- (orch);
\draw[arrow] (orch) -- (agent);
\draw[arrow] (agent.east) -- (ext.west);

\draw[arrow] (agent.south) -- (telemetry.north);

\draw[arrow] (telemetry.east) --
node[midway, above=2pt, fill=white, inner sep=1pt] {\scriptsize evaluation driven evolution}
(eval.west);

\draw[arrow] (eval.east) -- (registry.west);
\draw[arrow] (eval.south) --
node[midway, right=2pt, fill=white, inner sep=1pt] {\scriptsize scores + regressions}
(opt.north);

\draw[arrow] (registry.south) |- (opt.east);
\node[fill=white, inner sep=1pt]
  at ($(opt.east)!0.55!(registry.south) + (6mm,2mm)$) {\scriptsize candidate artifacts};

\draw[dashedarrow]
(opt.west) .. controls +(-2.2,0) and +(-2.2,-2.0) ..
node[left, fill=white, inner sep=1pt] {\scriptsize gated deploy / rollback}
(agent.west);

\coordinate (divY) at ($(agent.south)+(0,-6mm)$);
\coordinate (divL) at ($(agent.west)+(-25mm,0)$);
\coordinate (divR) at ($(registry.east)+(8mm,0)$);

\draw[line width=0.6pt]
  ($(divL |- divY)$) -- ($(divR |- divY)$);

\node[anchor=west, fill=white, inner sep=1pt] at ($(divL |- divY)+(0,4mm)$) {\textbf{Execution Plane}};
\node[anchor=west, fill=white, inner sep=1pt] at ($(divL |- divY)+(0,-4mm)$) {\textbf{Control Plane (Self-Evolution)}};

\end{tikzpicture}%
}
\caption{Production architecture for a self-evolving support agent. Optimization and Safe Rollout consumes evaluator scores and regression signals together with candidate versioned artifacts, then emits gated deployment or rollback decisions to the execution plane.}
\label{fig:self-evolving-arch}
\end{figure*}
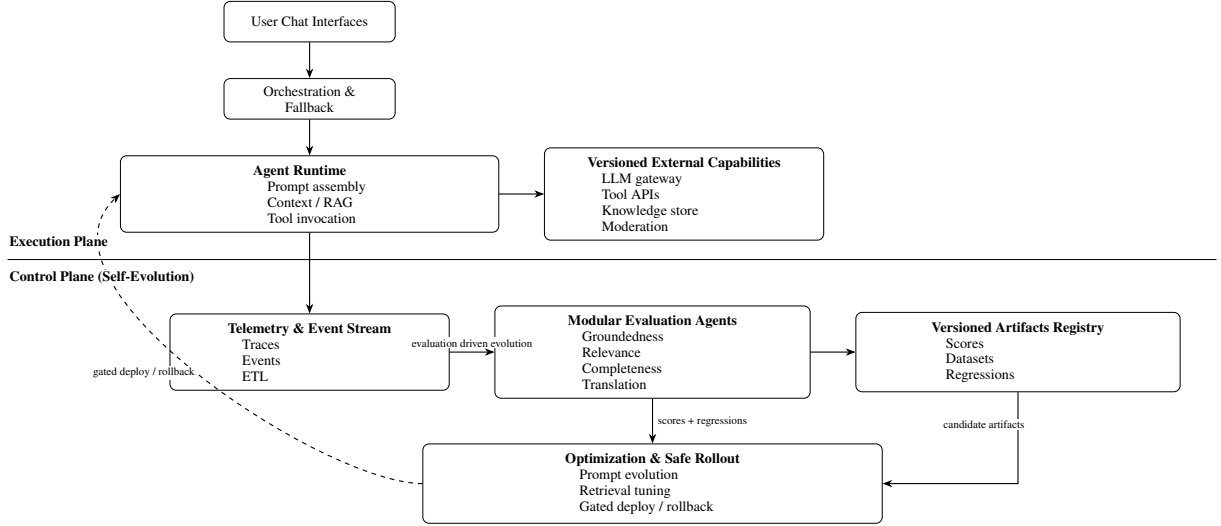

Continuous evolution of prompts, retrieval, and evaluation expands the failure surface relative to static deployments. The production architecture (Figure~\ref{fig:self-evolving-arch}) is structured around four principles.

\paragraph{Reliability as a prerequisite for self-evolution.}
A lightweight orchestration and fallback layer mediates user traffic with deterministic routing, error handling, and graceful degradation. This isolates end users from rapid iteration in the agent layer and enables fallback to human-assisted support when downstream AI components degrade or fail.

\paragraph{Runtime as a controlled mutation boundary.}
Rather than embedding behavior in code, prompts are assembled dynamically, context-construction policies are swappable, and tool usage is declaratively configured. Prompt evolution, retrieval tuning, and workflow updates therefore ship through configuration and artifact updates rather than code deploys, allowing behavioral improvements to propagate without redeploying the runtime itself.

\paragraph{Memory as versioned artifacts, not latent state.}
The runtime carries no cross-session latent state beyond short-horizon, task-scoped context required for conversational continuity. Longer-lived knowledge is promoted only through explicitly versioned and evaluated artifacts, enabling auditability, reproducibility, controlled rollback, and principled promotion of improvements while preventing uncontrolled memory drift.

\paragraph{Telemetry as an evaluation substrate.}
Execution telemetry serves both monitoring and automated evaluation. Structured event streams feed nearline quality checks and offline evaluators measuring groundedness, relevance, completeness, and policy compliance, forming the control plane that drives prompt evolution, retrieval tuning, and gated deployment without direct model retraining.

\paragraph{Optimization cadence.}
The outer optimization loop typically runs weekly and can also be triggered by content refreshes or detected regressions. One end-to-end evolution cycle takes hours to days depending on dataset size and fitness-evaluation budget. Candidate prompts and retrieval configurations are promoted only after offline regression checks and staged rollout; the execution plane can roll back to the prior versioned artifact if guardrails fail.

\paragraph{Deployment learnings.}
In production, three patterns proved load-bearing. \emph{Centralized execution} through configuration, modular prompts, and declarative tool selection---rather than per-LoB code forks, or separately maintained code paths for individual lines of business---was critical to avoiding behavioral divergence and ensuring improvements generalized across workflows. \emph{Layered fail-safes} such as targeted query redirection allowed traffic to route to a known-safe content pool when a polluted vector index required hours-long reindexing, cutting recovery from hours to minutes and limiting manual intervention. Finally, \emph{staged rollout} of new execution patterns in greenfield workflows---new, low-traffic workflows used as controlled proving grounds---before extension to mature traffic let evolution mechanisms and evaluation signals mature before broader deployment. Together these practices sustain continuous improvement under rapidly changing models, policies, and knowledge.

\section{Simulation and Ablation Studies}
\label{sec:simulation}
We evaluate the agent through controlled simulations that replay anonymized historical support interactions and synthetic edge cases across multilingual, product-diverse, and evolving knowledge settings. Each simulation acts as a system-level ablation, comparing configurations with different agent capabilities enabled under identical evaluation criteria.

\subsection{RAG Simulation}
\label{sec:rag-sim}
We compare four response-generation configurations using an LLM-based evaluator (GPT-4.1 via Azure OpenAI service): Vanilla RAG, RAG-only agents (ReAct-style and OpenAI tool invocation\footnote{OpenAI tool agent is accessed via Azure OpenAI service}), and the full Support AI Agent. The production generator is GPT-4o-mini while the offline judge is GPT-4.1, so they are different models. We measure overall quality, hallucination rate, and completeness with respect to retrieved support articles; a response is hallucinated if it contains any factual claim not attributable to the retrieved content. Table~\ref{tab:eval_ablation} calibrates the judge against human labels.

As shown in Table~\ref{tab:rag_simulation}, vanilla and agentic RAG configurations hallucinate at 4.8--6.2\%, while the full Support AI Agent reaches $<$0.1\% with the highest overall score (2.78) and completeness (87.8\%). The gain comes from explicit agentic constraints that prioritize synthesis from retrieved authoritative content and suppress reliance on parametric knowledge when grounded sources exist.

\begin{table*}[t]
\centering
\begin{threeparttable}
    \captionsetup{justification=centering}
    \caption{RAG Simulation Results}
    \label{tab:rag_simulation}

    \begin{tabular}{lccc}
    \toprule
    \textbf{Setting\tnote{1}} & \textbf{Overall Score\tnote{2}} & \textbf{Hallucination\tnote{3}} & \textbf{Completeness\tnote{4}} \\
    \midrule
    Vanilla RAG                  & 2.59           & 5.3\%          & 78.7\% \\
    RAG Only (ReAct Agent)       & 2.56           & 4.8\%          & 79.0\% \\
    RAG Only (OpenAI Tool Agent) & 2.48           & 6.2\%          & 75.0\% \\
    Support AI Agent             & \textbf{2.78}  & \textbf{<0.1\%} & \textbf{87.8\%} \\
    \bottomrule
    \end{tabular}

    \begin{tablenotes}
        \small
        \item[1] The simulation uses an anonymized, human-validated set of 100 support interactions with multiple inference runs per interaction.
        \item[2] Overall score is on a 0--3 scale to measure the overall quality of the response.
        \item[3] A response is labeled as hallucinated if it contains any factual claim that cannot be attributed to the retrieved support articles provided to the model at inference time.
        \item[4] A response is labeled as complete if it fully addresses all aspects of the user inquiry.
    \end{tablenotes}
\end{threeparttable}
\end{table*}

\subsection{Auto-Prompt Simulation}
\label{sec:autoprompt-sim}
On the routing intent detection dataset, we iteratively optimize LLM-generated initial prompts across generations using the genetic algorithm pipeline (Algorithm~\ref{alg:genetic_prompt_optimization}). Table~\ref{tab:auto_prompt_simulation} shows that prompt accuracy improves consistently only when both crossover and mutation are enabled---increasing average accuracy from 62.6\% to 68.0\% after two generations, with the best prompt reaching 73.3\%. Removing either operator slows improvement and saturates early, indicating that prompt \emph{evolution}, not random sampling, is what drives sustained gains.

\begin{table*}[t]
\centering
\begin{threeparttable}
    \captionsetup{justification=centering}
    \caption{Automatic Prompt Optimization Simulation}
    \label{tab:auto_prompt_simulation}

    \begin{tabular}{lccc}
    \toprule
    \textbf{Generation} & \textbf{Crossover + Mutation} & \textbf{No Mutation} & \textbf{No Crossover} \\
    \midrule
    0 (Baseline) & 62.6 / 66.7 & 62.6 / 66.7 & 62.6 / 66.7 \\
    1            & 66.7 / 73.3 & 63.3 / 66.7 & 62.6 / 66.7 \\
    2            & 68.0 / 73.3 & 64.0 / 66.7 & 63.3 / 66.7 \\
    \bottomrule
    \end{tabular}

    \begin{tablenotes}
        \small
        \item The simulation uses a curated, human-validated intent-detection dataset (Search Knowledge Base vs Direct to Consultant; $N=30$). Metrics are average prompt accuracy / best prompt accuracy per generation.
    \end{tablenotes}
\end{threeparttable}
\end{table*}

\subsection{Evaluator-Signal Ablation}
\label{sec:eval-ablation}
Beyond the system-level simulations above, we ablate individual evaluator signals to isolate their contribution, replaying each configuration on an anonymized validation set and comparing against a human-labeled reference. As shown in Table~\ref{tab:eval_ablation}, response groundedness is the dominant driver of evaluator reliability---removing it causes the largest drop in alignment with human labels, while removing completeness barely changes it---and collapsing all signals into a single overall score performs worst, obscuring critical failure modes. These findings justify retaining groundedness, relevance, and completeness as first-class, independently reported dimensions~\cite{qiao2025worfbench,park2025mirage,es2023ragas}.

\begin{table}[!t]
\centering
\small
\setlength{\tabcolsep}{4pt}
\renewcommand{\arraystretch}{0.95}
\begin{threeparttable}
    \captionsetup{justification=centering}
    \caption{Ablation Study of RAG Auto-Evaluation Dimensions}
    \label{tab:eval_ablation}

    \begin{tabular}{l c}
    \toprule
    \textbf{Evaluation Setting} & \makecell[c]{\textbf{Alignment\tnote{1}}\\\textbf{w/ Human Labels\tnote{2}}} \\
    \midrule
    Full RAG Evaluation (Baseline) & \textbf{87\%} \\
    \midrule
    A1: No Groundedness       & 76\% (-11) \\
    A2: No Content Relevance  & 82\% (-5) \\
    A3: No Completeness       & 86\% (-1) \\
    A4: Overall Score Only    & 67\% (-20) \\
    \bottomrule
    \end{tabular}

    \begin{tablenotes}
        \small
        \item[1] Alignment is measured against human-labeled ground truth on a randomly sampled set of 100 premium-tier subscription support chats, with multiple runs per interaction.
        \item[2] Three trained reviewers, blind to system condition, followed standardized guidelines. Raw agreement was 92\%; disagreements were resolved by majority vote.
    \end{tablenotes}
\end{threeparttable}
\end{table}

\subsection{Multilingual Translation Evaluation}
\label{sec:translation-eval}
This evaluator-design ablation tests the modular multi-agent evaluator that produces the fitness signals driving Auto-Prompt and RAG iteration. We compare it on the En--Zh slice with COMET~\cite{rei2020comet} and a single-agent LLM judge. As summarized in Table~\ref{tab:translation_eval_simulation}, accuracy rises from 49.7\% (COMET) to 76.5\% (single-agent LLM judge) to 84.8\% (modular multi-agent evaluator): decomposing translation quality into specialized dimensions captures the domain-specific, context-sensitive demands of support interactions that metric-based scoring misses.

\begin{table}[t]
\centering
\small
\begin{threeparttable}
    \captionsetup{justification=centering}
    \caption{Translation Quality Evaluation Result\tnote{1}}
    \label{tab:translation_eval_simulation}

    \begin{tabular}{l c}
    \toprule
    \textbf{Evaluation Paradigm} & \textbf{Accuracy (\%)} \\
    \midrule
    Model-based eval (COMET)   & 49.7 \\
    Pure LLM single-agent eval & 76.5 \\
    Modular multi-agent eval   & \textbf{84.8} \\
    \bottomrule
    \end{tabular}

    \begin{tablenotes}
        \small
        \item[1] Evaluation is performed on an anonymized support chat dataset (N=300).
    \end{tablenotes}
\end{threeparttable}
\end{table}

\section{Online Experiment}
We ran a two-week A/B test on continuous live member and customer support traffic. Users were randomized once with fixed 50/50 assignment and remained in one arm for the experiment, preventing repeated-user contamination across conditions. The control was the original production agent---a handcrafted prompt, fixed retrieval pipeline, and periodic human QA. The treatment was the integrated self-evolved workflow, which bundles (i) evolved Auto-Prompt, (ii) agent-invoked RAG, and (iii) closed-loop evaluator-driven iteration with gated rollout. The three metrics use separate randomized subpopulations.

The experiment was planned for four weeks with sequential monitoring and concluded after two weeks, when the effects were stable, power exceeded 99.9\%, and significance remained under alpha-spending corrections. We compare each proportion with a two-sided two-proportion $z$-test (equivalently, a $\chi^2$ test of independence) and apply Holm correction across the three primary outcomes. The user is the inference unit; average conversations per user are approximately 1.26 for QA and 1.53 for cancellation. The corresponding cluster adjustment is modest (design effect $\leq 1.5$), and user-clustered GEE/CR2 standard errors preserve significance for all outcomes; for example, the cancellation statistic changes from $z=10.0$ to $z=8.1$. Table~\ref{tab:online_experiment_results} reports the unclustered point estimates and confidence intervals.

\paragraph{QA self-serve.} This is the share of product-question or technical conversations resolved without human escalation. It increased from 33.7\% to 42.7\%, an absolute lift of 9.0 percentage points (95\% CI [8.4, 9.6]; $z=27.6$).

\paragraph{Cancellation self-serve.} This is the share of cancellation-intent conversations completed end-to-end without handoff. It increased from 61.9\% to 66.6\%, an absolute lift of 4.8 percentage points (95\% CI [3.8, 5.7]; $z=10.0$).

\paragraph{Routing accuracy.} This is the share routed to the correct human queue against the labeled target. It increased from 38.2\% to 68.8\%, an absolute lift of 30.6 percentage points (95\% CI [23.6, 37.6]; $z=8.2$).

All three outcomes have $p\ll10^{-4}$ and remain significant after Holm correction. We also monitored escalation rate, thumbs-up/down feedback, latency, customer-satisfaction score, and moderation incidents; none regressed under treatment. The online study reports the end-to-end system effect of the integrated self-evolved workflow, while Tables~\ref{tab:rag_simulation}--\ref{tab:eval_ablation} isolate the components offline.

\begin{table*}[t]
\centering
\begin{threeparttable}
    \captionsetup{justification=centering}
    \caption{Online Experiment Results}
    \label{tab:online_experiment_results}

    \small
    \setlength{\tabcolsep}{3pt}
    \begin{tabular}{lcccc}
    \toprule
    \textbf{Metric} & \makecell{\textbf{Control}\\original production\\agent} & \makecell{\textbf{Treatment}\\integrated self-evolved\\workflow} & \makecell{\textbf{Absolute Lift}\\\textbf{(95\% CI)}} & \textbf{$z$} \\
    \midrule
    QA self-serve\tnote{1}           & 33.7\% & 42.7\% & +9.0 pp [8.4, 9.6] & 27.6 \\
    Cancellation self-serve\tnote{2} & 61.9\% & 66.6\% & +4.8 pp [3.8, 5.7] & 10.0 \\
    Routing accuracy\tnote{3}        & 38.2\% & 68.8\% & +30.6 pp [23.6, 37.6] & 8.2 \\
    \bottomrule
    \end{tabular}

    \begin{tablenotes}
        \footnotesize
        \item[1] QA: 42,982 / 45,669 conversations and 35,867 / 34,334 users (control / treatment).
        \item[2] Cancellation: 20,468 / 20,456 conversations and 13,442 / 13,308 users (control / treatment).
        \item[3] Routing: 356 / 356 decisions (control / treatment) on a fixed labeled evaluation set.
        \item[] Two-sided two-proportion $z$-tests; all $p\ll10^{-4}$ and significant after Holm correction.
    \end{tablenotes}
\end{threeparttable}
\end{table*}

Beyond aggregate gains, these results show that the integrated self-evolved workflow improved the measured outcomes over the two-week deployment window, consistent with calls for holistic, production-oriented evaluation~\cite{liang2023helm,es2023ragas,qiao2025worfbench}.

\section*{Limitations}

\paragraph{Evaluator dependence.} The closed loop relies on LLM-as-judge evaluators (GPT-4.1) for fitness signals. Our ablations measure alignment with human labels (Table~\ref{tab:eval_ablation}); larger samples and cross-checks with open evaluators would further characterize variance and model-specific judge sensitivity~\cite{zheng2023mtbench,dubois2024lengthcontrolled}.

\paragraph{Cost and latency of evolution.} Genetic prompt search incurs additional inference cost per generation, and end-to-end evolution cycles take hours to days depending on dataset size and fitness budget. Future work could study tighter adaptation windows for rapidly changing knowledge.

\paragraph{Single-tenant evaluation.} The online study covers two weeks on one deployment surface at a single enterprise. Longer observation windows and deployments in other domains would test durability and transfer to different tool inventories and operating environments.

\paragraph{Online component attribution.} The online treatment bundles Auto-Prompt, agent-invoked RAG, and evaluator-driven iteration as the integrated self-evolved workflow. Factorial deployments could quantify the marginal contribution and interaction of each component.

\paragraph{Retrieval ceiling.} Response quality is bounded by retrieval coverage. When the underlying content lake lacks an authoritative document, the agent's grounding constraints correctly suppress hallucination but cannot synthesize a correct answer; further gains require improvements to retrieval recall and authoritative-content coverage that are orthogonal to the self-evolution loop.

\paragraph{Limited multilingual stress test.} The controlled translation study focuses on English$\leftrightarrow$Chinese (Section~\ref{sec:translation-eval}). Systematic evaluation of low-resource languages and code-mixed input remains an important extension.

\paragraph{Closed-source dependencies.} The production system depends on closed-source foundation models (GPT-4o-mini, GPT-4.1) and a proprietary search backend. Reproduction with open-source equivalents (e.g., Llama 3, BGE retrievers) is feasible in principle but has not been verified end-to-end.

\section*{Acknowledgments}

\makeatletter
\ifacl@anonymize\else
We thank Artem Grigoryan, Tony Huynh, Zhentao Lin, Chris Korbel, Umang Lahoti and Ajay Vishwanathan for their support and collaboration on the agent application and A/B test.
\fi
\makeatother

\paragraph{Use of AI assistance.} In preparing this manuscript, the authors used Anthropic's Claude (via Claude Code) for prose polishing, LaTeX formatting and table layout, restructuring of section ordering, and drafting assistance on the Limitations section. All technical claims, experimental design, results, figures, and citations were authored, reviewed, and verified by the human authors, who take full responsibility for the content of the paper.

\bibliography{custom}

@inproceedings{zhou2023ape,
  title     = {Large Language Models Are Human-Level Prompt Engineers},
  author    = {Zhou, Yongchao and others},
  booktitle = {ICLR},
  year      = {2023}
}

@inproceedings{yang2024opro,
  title     = {Large Language Models as Optimizers},
  author    = {Yang, Chengrun and others},
  booktitle = {ICLR},
  year      = {2024}
}

@inproceedings{guo2024evoprompt,
  title     = {Connecting Large Language Models with Evolutionary Algorithms Yields Powerful Prompt Optimizers},
  author    = {Guo, Qingyan and others},
  booktitle = {ICLR},
  year      = {2024}
}

@misc{fernando2024promptbreeder,
  title  = {Promptbreeder: Self-Referential Self-Improvement via Prompt Evolution},
  author = {Fernando, Chrisantha and others},
  year   = {2023},
  note   = {arXiv:2309.16797}
}

@misc{khattab2023dspy,
  title     = {DSPy: Compiling Declarative Language Model Calls into Self-Improving Pipelines},
  author    = {Khattab, Omar and others},
  year      = {2023},
  note      = {arXiv:2310.03714}
}

@article{zheng2023mtbench,
  title   = {Judging LLM-as-a-Judge with MT-Bench and Chatbot Arena},
  author  = {Zheng, Lianmin and others},
  journal = {arXiv:2306.05685},
  year    = {2023}
}

@inproceedings{liu2024geval,
  title     = {G-Eval: NLG Evaluation using GPT-4 with Better Human Alignment},
  author    = {Liu, Yang and others},
  booktitle = {EMNLP},
  year      = {2023}
}

@article{dubois2024lengthcontrolled,
  title   = {Length-Controlled AlpacaEval: A Simple Way to Debias Automatic Evaluators},
  author  = {Dubois, Yann and others},
  journal = {arXiv:2404.04475},
  year    = {2024}
}

@inproceedings{kim2024prometheus2,
  title     = {Prometheus 2: An Open Source Language Model Specialized in Evaluating Other Language Models},
  author    = {Kim, Seungone and others},
  booktitle = {EMNLP},
  year      = {2024}
}

@inproceedings{lewis2020rag,
  title     = {Retrieval-Augmented Generation for Knowledge-Intensive NLP Tasks},
  author    = {Lewis, Patrick and others},
  booktitle = {NeurIPS},
  year      = {2020}
}

@inproceedings{karpukhin2020dpr,
  title     = {Dense Passage Retrieval for Open-Domain Question Answering},
  author    = {Karpukhin, Vladimir and others},
  booktitle = {EMNLP},
  year      = {2020}
}

@inproceedings{gao2023hyde,
  title     = {Precise Zero-Shot Dense Retrieval without Relevance Labels},
  author    = {Gao, Luyu and others},
  booktitle = {ACL},
  year      = {2023}
}

@inproceedings{asai2024selfrag,
  title     = {Self-RAG: Learning to Retrieve, Generate, and Critique through Self-Reflection},
  author    = {Asai, Akari and others},
  booktitle = {ICLR},
  journal   = {arXiv:2310.11511},
  year      = {2024}
}

@article{yan2024crag,
  title   = {Corrective Retrieval Augmented Generation},
  author  = {Yan, Shi-Qi and others},
  journal = {arXiv:2401.15884},
  year    = {2024}
}

@inproceedings{yao2023react,
  title     = {ReAct: Synergizing Reasoning and Acting in Language Models},
  author    = {Yao, Shunyu and others},
  booktitle = {ICLR},
  year      = {2023}
}

@article{ovadia2024finetuneorretrieve,
  title   = {Fine-Tuning or Retrieval? Comparing Knowledge Injection in LLMs},
  author  = {Ovadia, Oded and others},
  journal = {arXiv:2312.05934},
  year    = {2024}
}

@article{soudani2024finetuningvsrag,
  title   = {Fine Tuning vs. Retrieval Augmented Generation for Less Popular Knowledge},
  author  = {Soudani, Heydar and others},
  journal = {arXiv:2403.01432},
  year    = {2024}
}

@article{wu2023autogen,
  title   = {AutoGen: Enabling Next-Gen LLM Applications via Multi-Agent Conversation},
  author  = {Wu, Qingyun and others},
  journal = {arXiv:2308.08155},
  year    = {2023}
}

@article{packer2023memgpt,
  title   = {MemGPT: Towards LLMs as Operating Systems},
  author  = {Packer, Charles and others},
  journal = {arXiv:2310.08560},
  year    = {2023}
}

@article{shinn2023reflexion,
  title   = {Reflexion: Language Agents with Verbal Reinforcement Learning},
  author  = {Shinn, Noah and others},
  journal = {arXiv:2303.11366},
  year    = {2023}
}

@inproceedings{lewis2019mlqa,
  title     = {MLQA: Evaluating Cross-lingual Extractive Question Answering},
  author    = {Lewis, Patrick and others},
  booktitle = {Proceedings of the 58th Annual Meeting of the Association for Computational Linguistics},
  year      = {2020}
}

@article{longpre2021mkqa,
  title   = {MKQA: A Linguistically Diverse Benchmark for Multilingual Open Domain Question Answering},
  author  = {Longpre, Shayne and others},
  journal = {Transactions of the Association for Computational Linguistics},
  volume  = {9},
  pages   = {1389--1406},
  year    = {2021},
  url     = {https://aclanthology.org/2021.tacl-1.82}
}

@inproceedings{qiao2025worfbench,
  title     = {Benchmarking Agentic Workflow Generation},
  author    = {Qiao, Shuofei and others},
  booktitle = {ICLR},
  year      = {2025},
  note      = {arXiv:2410.07869}
}

@article{park2025mirage,
  title   = {MIRAGE: A Metric-Intensive Benchmark for Retrieval-Augmented Generation Evaluation},
  author  = {Park, Chanhee and others},
  journal = {Findings of NAACL},
  year    = {2025},
  note    = {arXiv:2504.17137}
}

@article{es2023ragas,
  title   = {Ragas: Automated Evaluation of Retrieval Augmented Generation},
  author  = {Es, Shahul and others},
  journal = {arXiv:2309.15217},
  year    = {2023}
}

@article{liang2023helm,
  title   = {Holistic Evaluation of Language Models},
  author  = {Liang, Percy and others},
  journal = {Transactions on Machine Learning Research (TMLR)},
  year    = {2023},
  note    = {arXiv:2211.09110}
}

@article{rei2020comet,
  title={COMET: A Neural Framework for MT Evaluation},
  author={Rei, Ricardo and others},
  journal={arXiv preprint arXiv:2009.09025},
  year={2020}
}

@inproceedings{lin2022xglm,
  title={Few-shot Learning with Multilingual Generative Language Models},
  author={Lin, Xi Victoria and others},
  booktitle={Proceedings of the 2022 Conference on Empirical Methods in Natural Language Processing (EMNLP)},
  pages={9019--9052},
  year={2022}
}

@book{holland1992adaptation,
  title={Adaptation in Natural and Artificial Systems},
  author={Holland, John H.},
  publisher={MIT Press},
  year={1992}
}

@article{whitley1994genetic,
  title={A Genetic Algorithm Tutorial},
  author={Whitley, Darrell},
  journal={Statistics and Computing},
  year={1994}
}

@article{guu2020realm,
  title={REALM: Retrieval-Augmented Language Model Pre-Training},
  author={Guu, Kelvin and others},
  journal={ICML},
  year={2020}
}

@article{schick2023toolformer,
  title={Toolformer: Language Models Can Teach Themselves to Use Tools},
  author={Schick, Timo and others},
  journal={NeurIPS},
  year={2023}
}

@article{lin2021pretrained,
  title={Pretrained Transformers for Text Ranking: BERT and Beyond},
  author={Lin, Jimmy and others},
  journal={Synthesis Lectures on Human Language Technologies},
  year={2021}
}

@article{gao2023rag,
  title={Retrieval-Augmented Generation for Large Language Models: A Survey},
  author={Gao, Yunfan and others},
  journal={arXiv:2312.10997},
  year={2023}
}

@article{nogueira2019passage,
  title={Passage Re-ranking with BERT},
  author={Nogueira, Rodrigo and Cho, Kyunghyun},
  journal={arXiv:1901.04085},
  year={2019}
}

@inproceedings{press2023iterative,
  title={Measuring and Narrowing the Compositionality Gap in Language Models},
  author={Press, Ofir and others},
  booktitle={Findings of EMNLP},
  year={2023},
  note={arXiv:2210.03350}
}

@online{microsoft2023hybrid,
    author = {Alec Berntson},
    title = {Azure AI Search: Outperforming vector search with hybrid retrieval and reranking},
    year = {2023},
    url = {https://techcommunity.microsoft.com/blog/azure-ai-foundry-blog/azure-ai-search-outperforming-vector-search-with-hybrid-retrieval-and-reranking/3929167},
    urldate = {2024-02-04},
    organization = {Microsoft Research}
}

\end{document}